\documentclass{article}

\usepackage{PRIMEarxiv}

\usepackage[utf8]{inputenc} 
\usepackage[T1]{fontenc}    
\usepackage{hyperref}       
\usepackage{url}            
\usepackage{booktabs}       
\usepackage{amsfonts}       
\usepackage{nicefrac}       
\usepackage{microtype}      
\usepackage{lipsum}
\usepackage{fancyhdr}       
\usepackage{graphicx}       
\graphicspath{{media/}}     

\title{KNOW--A Real-World Ontology for Knowledge Capture with Large Language Models}

\author{
  Arto Bendiken\\
  Haltia, Inc. \\
  \texttt{arto@haltia.ai} \\
}
\begin{document}
\maketitle

\begin{abstract}
We present KNOW--the Knowledge Navigator Ontology for the World--the first ontology designed to capture everyday knowledge to augment large language models (LLMs) in real-world generative AI use cases such as personal AI assistants. Our domain is human life, both its everyday concerns and its major milestones. We have limited the initial scope of the modeled concepts to only established human universals: spacetime (places, events) plus social (people, groups, organizations). The inclusion criteria for modeled concepts are pragmatic, beginning with universality and utility. We compare and contrast previous work such as Schema.org and Cyc--as well as attempts at a synthesis of knowledge graphs and language models--noting how LLMs already encode internally much of the commonsense tacit knowledge that took decades to capture in the Cyc project. We also make available code-generated software libraries for the 12 most popular programming languages, enabling the direct use of ontology concepts in software engineering. We emphasize simplicity and developer experience in promoting AI interoperability.
\end{abstract}


\section{Introduction}
While not yet widely known in industry, the neuro-symbolic approach to artificial intelligence (AI)--in particular, the union of large language models (LLMs) and knowledge graphs (KGs)--has in academia already established itself as the most promising state-of-the-art pathway towards building practical, trustworthy, explainable, and interoperable AI systems. We will briefly outline the promise of the neuro-symbolic synthesis and explain why common ontologies are needed for reliability and interoperability.

\subsection{Why LLMs Need KGs}
Despite limited successes with prototypes which escaped the lab, over the course of the next several years industry attempts to deploy standalone LLMs in production will be increasingly frustrated by the limitations and deficiencies of said models: the limited context window and its poor scaling characteristics, the lack of introspectability and justifiability, the propensity towards hallucinations, as well as ultimately the implicit and static nature of the very knowledge encoded during pre-training and the difficulty of amending it afterwards.

These various problems can be mitigated or solved by a hybrid approach where the LLM accesses and manipulates a symbolic knowledge base where facts are captured and represented in explicit form.

\subsection{Why KGs Need LLMs}
Conversely, large language models do already encode internally much of the commonsense tacit knowledge that in previous decades proved the most formidable challenge to symbolic approaches of knowledge representation and AI. Until recently, any toddler ultimately knew more about the nature and causal structure of our physical and social worlds than any computer. The ambition of the decades-long Cyc project \cite{lenat_building_1989} that sought to capture and encode all commonsense knowledge was amply matched by the interminability of the endeavor.

An LLM's “knowledge soup”--a large reservoir of loosely organized encyclopedic knowledge \cite{sowa_crystallizing_1990}--is quantitatively derived from statistical distributions in an LLM’s pre-training dataset and after training is implicitly encoded in the billions of weights that constitute the model \cite{templeton2024scaling}. As mentioned, it is not particularly introspectable nor is it even consistent, but it is nonetheless useful: it largely sidesteps the need to explicitly represent and encode truly basic tenets of the world, such as that the noon sky is blue, that the arrow of gravity points downwards, and that humans tend to live in families and build housing to seek shelter from the elements.

\subsection{The Neuro-Symbolic Synthesis}
In this way, neural networks in the form of LLMs provide the missing ingredient for a hybrid that realizes long-standing visions for symbolic AI beyond narrow expert system use cases. And conversely, the explicit knowledge representation in KGs complements and enables LLMs to transcend their limitations and ultimately be safely deployed in real-world use cases beyond technology demos.

To realize the promise of this neuro-symbolic synthesis, however, the language model must understand the restricted vocabulary (that is, the set of symbols) used for knowledge representation. This is where the need for an ontology comes into play. The ontology restricts and guides knowledge capture \cite{coplu_prompt-time_2024}, defines and describes the semantics of what has been captured \cite{allemang_increasing_2024}, and enables the LLM to give better, more precise responses.

We present herein the first iteration of our ongoing effort to design a commonsense, everyday ontology suitable for real-world use cases for neuro-symbolic AI. We have appropriately titled our ontology KNOW, a backronym standing for Knowledge Navigator Ontology for the World. Our ontology is open source--indeed, it is placed into the public domain with no strings attached.

\section{Domain and Scope}

\begin{figure}[ht]
  \centering
  \includegraphics[scale=0.35]{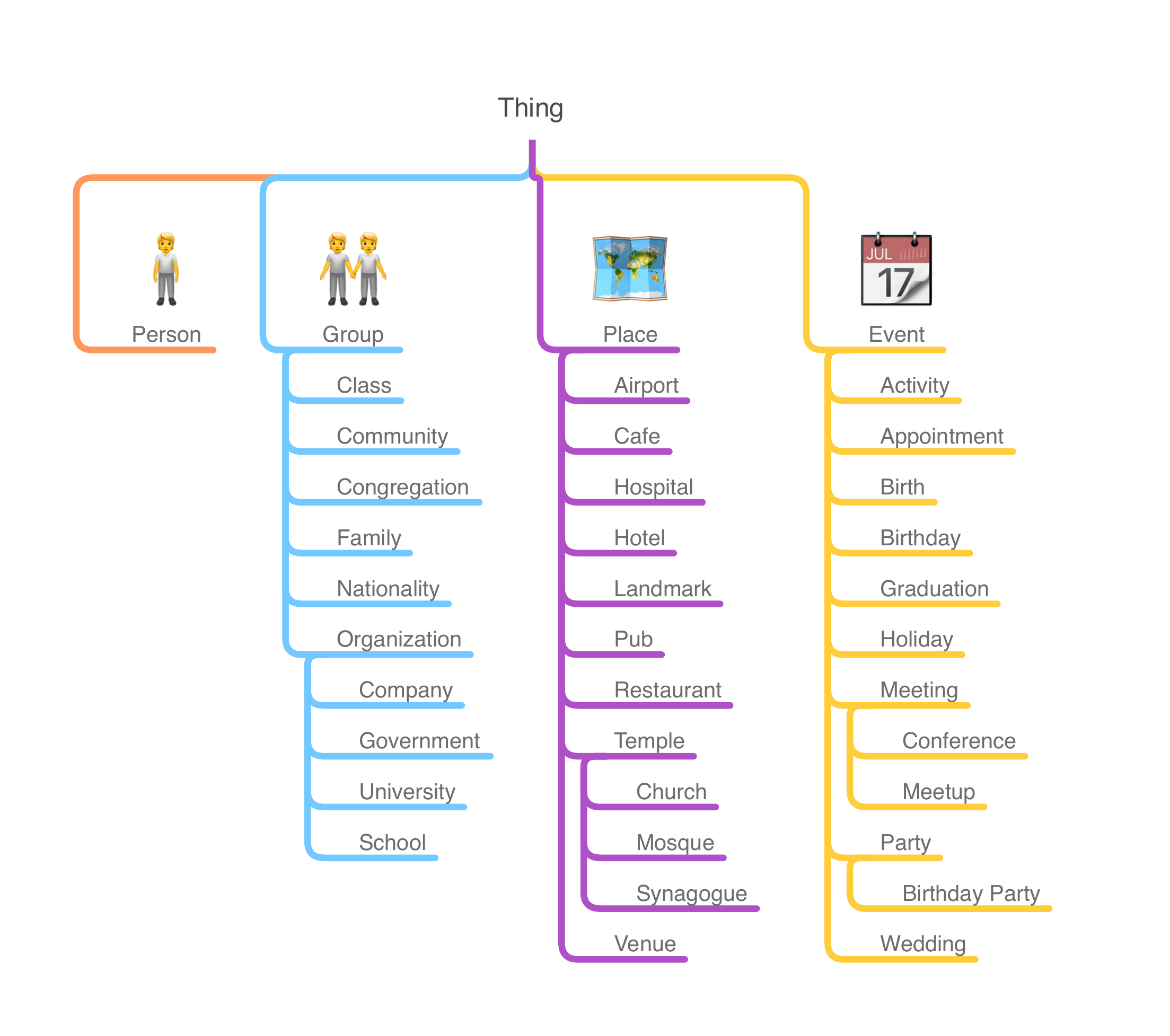}
  \caption{Top-level classes in the ontology (not comprehensive)}
  \label{fig:fig1}
\end{figure}

Our domain is human life, both its everyday concerns and its major milestones. We have limited the initial scope of the ontology to only established human universals: spacetime (places, events) plus social (people, groups, organizations). Top-level classes are illustrated in Figure \ref{fig:fig1}. We anticipate extending this scope in future work.

Human universals “\textit{comprise those features of culture, society, language, behavior, and psyche for which there are no known exception}” \cite{brown_human_1991}. That is, human universals are common to all human societies ever studied. They include, for example, language, gossip, jokes, toys, tools, promises, symbolism, semantic categories, and logical notions.

We as humans are embedded in spacetime. To understand and interact with the physical world in which we live, time and space are the fundamental concepts that provide the basic framework for navigating and orienting ourselves in the world as well as the dimensions in which all physical events occur. From spacetime, human universals, and language essentials, we derive the base concepts \texttt{Place} and \texttt{Event}.

Examples of places common to all human societies include landmarks and places of worship. For contemporary developed societies, we also include airports, cafes, hospitals, hotels, and restaurants. Some examples of events for the same include birthdays, appointments, holidays, meetings, and parties.

As highly social creatures, the stature of the social reality--people and relationships--that we inhabit is nearly that of spacetime. Indeed, people have been known to prioritize consensus social reality over objective physical reality in their understanding of the world and actions in it. In this social dimension, then, we include the base concepts of \texttt{Person} and \texttt{Group}, the latter denoting a group of people. Unlike a biologist’s classification of a person as a human, hominid, anthropoid, primate, mammal, and vertebrate animal, we are here concerned only with commonsense understanding\footnote{Indeed, the distinction here is perhaps most clearly illustrated by considering the question of animals: scientific taxa and evolutionary lineages aside, from an early age children tend to consider themselves as something categorically distinct from animals.} and omit these taxonomic superclasses--at least in this first formulation of the ontology.

\section{Use Cases}
Our primary use cases revolve around everyday human concerns. For example, and in particular:

\textbf{Social relationships and roles.} Family, friends, acquaintances, coworkers, authority figures, and the various roles people play in society.

\textbf{Basic human needs.} Shelter, heating, cooling, water, food, healthcare, safety, etc. \cite{gupta_2008}

\textbf{Emotions and mental states.} Happiness, sadness, anger, fear, love, desire, beliefs, intentions, etc.

\textbf{Daily activities and routines.} Work, school, household chores, leisure activities, transportation, etc.

\textbf{Time and events.} Temporal concepts (past, present, future, duration, etc.), and common events and life milestones that occur in human life (birth, education, employment, marriage, illness, death, etc.)

For secondary use cases, we also wish to extend support to fictional uses, especially modeling fictional social graphs--including in speculative fiction. This means, for example, that we don’t constrain the value range of the \texttt{age} property on \texttt{Person}. And on the other hand, we call the class \texttt{Person} instead of \texttt{Human}; on the gripping hand, however, we haven’t at present included an \texttt{Agent} class, as that would veer away from practicability. Similarly, \texttt{Place} isn’t inherently something constrained only to Earth’s surface, though the supermajority of uses will be exactly that. When practical uses and fictional uses are in tension, we have aimed to make the fictional use possible yet the practical use convenient.

\section{Software Libraries}
Large language models have proven capable code generation tools, and are deployed both for advanced code completion in code editors as well as in more ambitious prompt-driven uses. Given how code synthesis is seeing an unprecedented surge in interest, we have explicitly designed the ontology to be usable towards this important end. This affects, for example, property naming, which is designed to accord with naming conventions for coding.

We have made available code-generated open-source software development kits (SDKs) for the 12 most popular programming languages, enabling the direct use of ontology concepts in software design and implementation. The initial set of programming languages comprises C, C++, C\#\footnote{And other programming languages on the Microsoft .NET platform.}, Dart, Go, Java\footnote{And other programming languages on the Java Virtual Machine (JVM) platform.}, JavaScript, Python, Ruby, Rust, Swift, and TypeScript.

Our primary objective for the SDKs is to be able to represent ontology concepts using the native composite data abstraction for each language. For example, we represent ontology classes as structures in C, Go, and Rust; and as classes in Dart, Java, Python, and Ruby.

Where possible (when supported by the language), we also define abstract interfaces (C\#, Dart, Go, Java, TypeScript), traits (Rust), or protocols (Swift) for the same. These are in fact the highest-value uses of our SDKs, since they make possible semantic interoperability between third-party libraries that each can depend on our SDK so as to achieve type compatibility between themselves.

A secondary objective for the SDKs is to enable data interchange via serializations in JSON and RDF form, where appropriate for the language and when a predominant RDF framework is available: for example, we support Jackson and Jena for Java, and RDF.rb for Ruby.

It is our intention to continue, through open-source community contributions, to extend language support over time so as to cover at least the top 20 most popular programming languages and establish the initiative as a cornerstone for polyglot programming.

\section{Compare and Contrast}
We compare and contrast previous work including Schema.org and Cyc.

\subsection{Schema.org}
The previous ontology nearest to our endeavor is undoubtedly Schema.org \cite{guha_schemaorg_2016}, and we make sure to map our concepts (classes and properties) to corresponding Schema.org concepts where possible. Though there is inevitably some overlap, the two projects have different focuses and ultimately different audiences.

Whereas Schema.org grew out of microdata (structured data markup on web pages) and thus focuses on describing \textit{things} such as creative works, KNOW is specifically designed for neuro-symbolic uses and focuses on describing \textit{life}, beginning with social relationships.

For example, during the first decade of the Schema.org initiative, it seems nobody has as yet needed to describe family relationships in greater detail than parent/child and sibling. KNOW, on the other hand, directly includes properties for father/mother, brother/sister, uncle/aunt, nephew/niece, and so on--materializing these most important personal relationships explicitly, without the need for graph traversal nor entailment at the point of lookup.

KNOW also differs in its insistence on modeling the world in a commonsense, everyday manner. We are not concerned with “correct” taxonomy, but merely pragmatic taxonomy, leading to a relatively flat class hierarchy. A further difference is due to our support for code generation, described in a previous section.

\subsection{Cyc (OpenCyc, ResearchCyc)}
In seeking to model commonsense knowledge, our most esteemed predecessor is the already mentioned Cyc project. The differences, though, are also clear. For one, we don’t aspire to be an upper ontology; for another, the expressiveness of our ontology is merely the description logics of OWL, not the undecidable higher-order logic (HOL) used in Cyc.

Our initiative also uses open source instead of proprietary licensing, and a public and collaborative development model instead of a closed and internal one. Our development cycle is iterative--acknowledging that worse is better \cite{gabriel_1989}--and driven by real-world use cases.

\section{Availability and Usage}
The ontology is developed using an open-source ethos and a public, collaborative, and iterative development process at GitHub (\href{https://github.com/KnowOntology}{github.com/KnowOntology}), with documentation and downloads made available on a dedicated website (\href{https://know.dev}{know.dev}).

\subsection{Base IRI and Prefix}
The base IRI for the ontology is \texttt{https://know.dev/}, and our recommended prefix for CURIE construction is \texttt{know}. The base IRI has been designed to be memorable in support of our emphasis on developer experience. In Turtle format:

\begin{verbatim}
@base <https://know.dev/> .
@prefix know: <https://know.dev/> .
\end{verbatim}

\section{Conclusion}
We have outlined a first draft of KNOW, an ontology for capturing and representing everyday knowledge to augment LLMs in real-world generative AI use cases such as personal AI assistants, and explained how and where it fits into the emerging neuro-symbolic--or neuro-semantic, as it may be \cite{hermelen_2024}--synthesis.

It is worth keeping in mind the starting point that most practitioners working with LLMs today are as yet unaware of knowledge graphs, semantic technology, or even the need for an ontology for knowledge capture. Practitioners’ work generally proceeds, as it has for decades, based on ad-hoc properties conjured up on the spot when they were first needed, with little regard for stronger semantics or wider interoperability. This, though, will become increasingly untenable as an approach over the next several years.

While being based on a best-practice Semantic Web technology stack (that is, utilizing the RDF and OWL standards), by emphasizing wider availability and applicability, and prioritizing simplicity and the developer experience, we expect the KNOW initiative to serve as something of a Trojan horse for the dissemination of semantic technology in generative AI use cases more broadly as well. Much remains to be done, but the promise here is ultimately that of interoperable AI, where knowledge captured by one system can flow to and be utilized in other vendors’ systems as well.

\section*{Acknowledgements}
We would like to thank Tolga Çöplü and Stephen Cobb for their review and constructive feedback, which significantly enhanced the clarity and quality of this paper.

\bibliographystyle{apalike}
\bibliography{references}  

\end{document}